\documentclass[letterpaper, 10 pt, conference]{ieeeconf}
\IEEEoverridecommandlockouts                              

\usepackage{times}
\usepackage{epsfig}
\usepackage{graphicx}
\usepackage{amsmath}
\usepackage{amssymb}
\usepackage{graphicx}
\usepackage{amsmath}
\usepackage{booktabs}
\usepackage{algorithm}
\usepackage{algorithmic}
\usepackage{comment}
\usepackage{graphicx}
\usepackage{amsmath,amssymb} 
\usepackage{color}
\usepackage{booktabs}
\usepackage{bm}
\usepackage{makecell}

\usepackage[pagebackref=true,breaklinks=true,letterpaper=true,colorlinks,bookmarks=false]{hyperref}

\author{Jiakai Lin, Jinchang Zhang, Ge Jin, Wenzhan Song, Tianming Liu, Guoyu Lu
\thanks{Jiakai Lin, Jinchang Zhang and Guoyu Lu are with the Intelligent Vision and Sensing (IVS) Lab at SUNY Binghamton, USA. Ge Jin is with Yancheng Institute of Technology. Wenzhan Song and Tianming Liu are with the University of Georgia.
        {\tt\small guoyulu62@gmail.com}}%
}

\begin{document}

\title{3D Plant Root Skeleton Detection and Extraction}

\maketitle

\begin{abstract}
Plant roots typically exhibit a highly complex and dense architecture, incorporating numerous slender lateral roots and branches, which significantly hinders the precise capture and modeling of the entire root system. Additionally, roots often lack sufficient texture and color information, making it difficult to identify and track root traits using visual methods. Previous research on roots has been largely confined to 2D studies; however, exploring the 3D architecture of roots is crucial in botany. Since roots grow in real 3D space, 3D phenotypic information is more critical for studying genetic traits and their impact on root development.
We have introduced a 3D root skeleton extraction method that efficiently derives the 3D architecture of plant roots from a few images. This method includes the detection and matching of lateral roots, triangulation to extract the skeletal structure of lateral roots, and the integration of lateral and primary roots. We developed a highly complex root dataset and tested our method on it. The extracted 3D root skeletons showed considerable similarity to the ground truth, validating the effectiveness of the model.
This method can play a significant role in automated breeding robots. Through precise 3D root structure analysis, breeding robots can better identify plant phenotypic traits, especially root structure and growth patterns, helping practitioners select seeds with superior root systems. This automated approach not only improves breeding efficiency but also reduces manual intervention, making the breeding process more intelligent and efficient, thus advancing modern agriculture.

\end{abstract}

\section{Introduction}
\vspace{-1mm}
Roots serve as the primary channel through which plants interact with their physical and biological environments, playing a crucial role in nutrient absorption from the soil, which is essential for plant survival and growth \cite{khan2016root}\cite{rogers2015regulation}. Through 3D reconstruction technology, researchers can obtain precise models of root systems \cite{lu2021simultaneous}\cite{lu20213d}, providing valuable data for agriculturalists and biologists to measure characteristics such as root length, volumetric biomass, and overall crop development \cite{chen2016deep}\cite{chen2017cross}. However, the complex structure of root systems, with numerous slender branches that appear similar in shape and color, makes 3D reconstruction and feature identification highly challenging.
With advancements in deep learning and 3D reconstruction, researchers can now create 3D representations from single or multiple images \cite{fan2017point}\cite{choy20163D}\cite{wang2018pixel2mesh}\cite{yang20173D}, generating point clouds or volumetric representations \cite{choy20163D}\cite{yang20173D}\cite{fan2017point}\cite{wang2018pixel2mesh}, improving the accuracy of root system reconstruction. Nonetheless, due to the complexity and lack of texture in root systems, current methods still face challenges in extracting fine details and matching features, often resulting in low-density point clouds and incomplete models with occlusions and gaps. Additionally, improving accuracy typically requires multiple precisely calibrated cameras to capture images simultaneously, significantly increasing both cost and complexity.
In this context, utilizing robots for automated breeding has emerged as a new research direction. Robots can automatically capture images of root systems and perform 3D reconstruction, helping researchers precisely analyze root structure and health. This type of breeding robot, based on 3D root system reconstruction, can automatically select high-quality seeds by identifying plants with superior root structures and strong nutrient absorption capabilities. This improves breeding efficiency and reduces human intervention. Not only does it minimize reliance on manual observation in traditional breeding processes, but it also provides strong technical support for the automation and intelligent development of the breeding field.

This paper proposes a method for reconstructing the 3D skeleton of plant roots using a small number of multi-angle images. The 3D structure of roots is crucial for understanding plant phenotypes and for water and nutrient absorption. Accurate reconstruction can help breeders optimize root structures, improving crop yield and adaptability.
Our method consists of two parts: lateral root skeleton extraction and 3D spatial connections. First, we model the lateral roots and connect their starting points to the main root, building the topology of the entire root system. The spatial coordinates of the lateral roots are determined using triangulation from images taken at different angles. Before triangulation, we use a root detection network to mark the start and end points of the roots and identify the corresponding lateral root coordinates in two images. The feature extraction module selects the bounding box with the most matching keypoints as the corresponding box for the same lateral root. To minimize triangulation errors caused by detection network inaccuracies, we perform triangulation twice using images from different angles and optimize the results based on reprojection errors, matching corresponding coordinates and eliminating incorrect matches. The skeleton is further optimized through a Bundle Adjustment (BA) network, with skeleton points, root angles, and camera poses as inputs, minimizing reprojection errors and angular discrepancies between the roots. The lateral roots are connected from top to bottom, ultimately completing the extraction of the main root skeleton and achieving high-precision 3D reconstruction.

To the best of our knowledge, we believe we are the first to utilize images for the extraction of 3D root skeletons. Our contributions are summarized as follows: 
1. We have developed a lateral root matching network capable of identifying the positions and growth directions of lateral roots within root images, and detect and match the corresponding lateral roots from different views.
2. We have developed a method that can differentiate the primary roots and lateral roots, which can separately be reconstructed in 3D space. This can provide critical support for agriculture researchers for counting the number of each type of roots in a root system. 
3. By using only few images, we developed a method that can reconstruct the highly complicated root skeletons in 3D space directly.  
4. To reduce errors in the 3D skeleton obtained from multiple views, we added an SBA layer to the network for bundle adjustment. Based on the invariance of skeletal angles, we perform self-supervised training to optimize the skeleton structure and camera poses.

\begin{figure*}[t]
\begin{center}
\includegraphics[width=16cm, height=7cm]{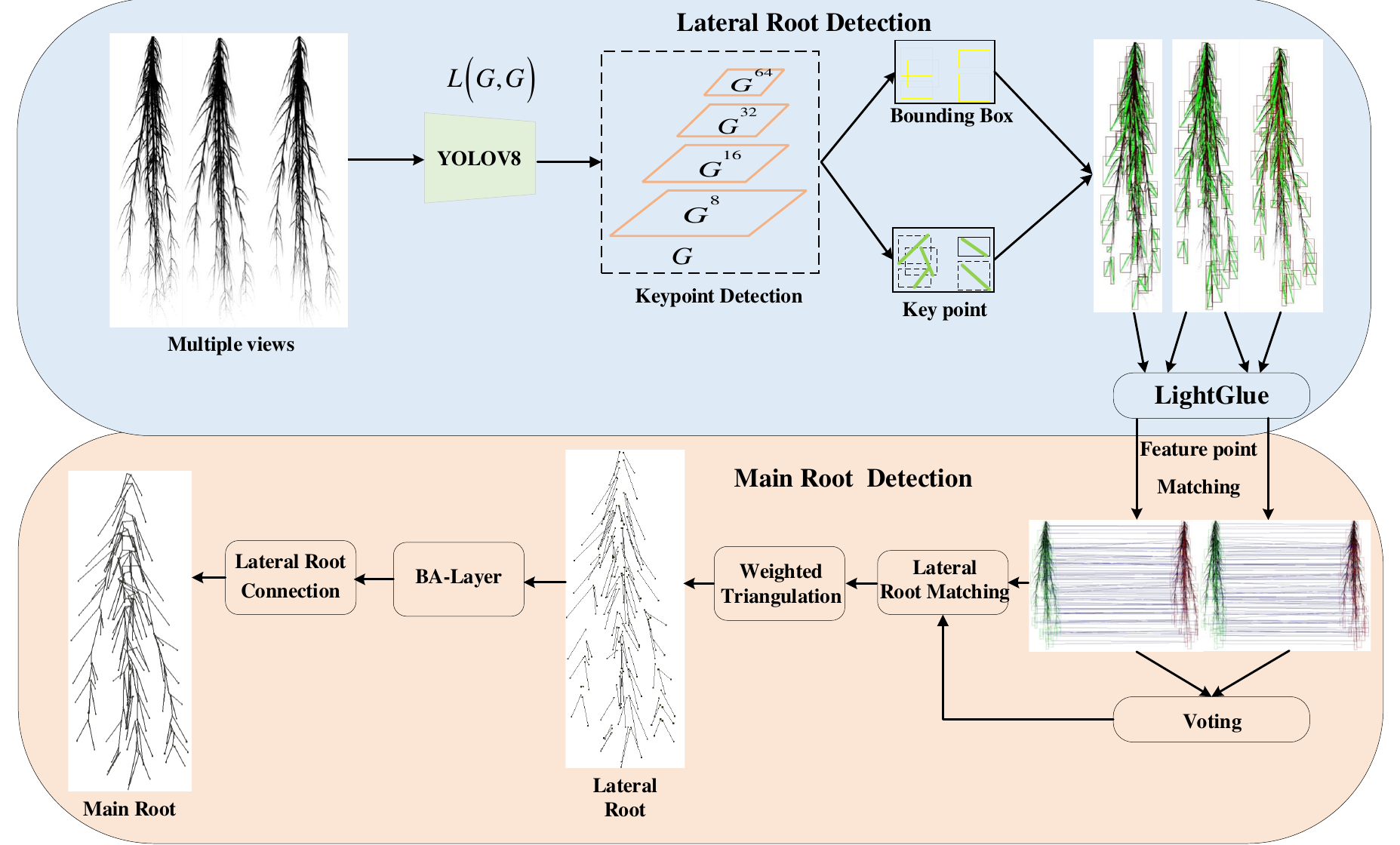}
\end{center}
\vspace{-5.5mm}
\caption{ Overview of our root keleton extraction  netwpork structure, which includes a detection network for identifying the positions and key points of each lateral root, a feature extraction network for extracting feature points from the lateral roots and finding matching feature points across multiple views, and a skeleton Bundle Adjustment (SBA) network for globally adjusting the skeletal structure and camera poses.}
\label{arch}
\vspace{-5.5mm}
\end{figure*}

\vspace{-1mm}
\section{RELATED WORK}
\vspace{-1mm}
\subsection{Multi-view Stereo}
\vspace{-1mm}
Initially, traditional algorithms such as Structure from Motion \cite{ozyecsil2017survey} and Simultaneous Localization and Mapping \cite{fuentes2015visual} used feature matching to map 2D pixels to 3D positions. Monocular depth estimation methods \cite{zhang2025vision}\cite{zhang2024embodiment} estimate the depth of overlapping roots, but all of these approaches tend to produce errors.
Currently, neural network algorithms \cite{huang2018deepmvs}\cite{ji2017surfacenet} have become the primary approach for multi-view reconstruction, indicating a significant improvement in handling complexity. CNN-based methods simultaneously extract features from each view and then combine these features into a unified representation of the object. Pooling-based fusion methods, such as those in \cite{liu2019simple} merge feature maps from different views and reduce their size through maximum or average pooling layers. Although simple, this method is less effective due to the lack of adjustable parameters. To capture the intricate latent relationships between views, transformer-based models have been introduced. 
\vspace{-1mm}
\subsection{3D Skeleton Detection}
\vspace{-1mm}
In spatio-temporal modeling, many CNN-based methods convert 3D skeleton sequences into pseudo-image formats to facilitate feature learning \cite{ding2017investigation}\cite{xu2018ensemble}. For example, \cite{wang2016action} introduced Joint Trajectory Maps (JTM), which use color encoding to convert the spatial configurations and dynamics of joint trajectories into texture images. Though effective, this method results in some information loss. To address this, \cite{li2017skeleton} proposed a translation- and scale-invariant mapping strategy, dividing skeleton joints into five parts and mapping them into a 2D format, combining spatial and temporal information. 
While key point detection has advanced \cite{mcnally2022rethinking}, detecting the starting and ending points of primary and lateral roots remains challenging. 
Some researchers have applied computer vision to robotic breeding, enabling automatic plant trait analysis and improving efficiency, such as selecting high-quality seeds through deep learning.  Meanwhile, researchers have also been exploring lower-level visual modeling approaches. For example, \cite{lin2025keypoint} performs keypoint detection directly from raw Bayer images, demonstrating strong potential in low-level perceptual precision. On the other hand, ground-penetrating radar (GPR) can be deployed \cite{lu2025non}\cite{zhang2024underground} for underground target detection, offering an alternative perspective for tasks such as plant root modeling.
However, the complexity of root reconstruction presents significant challenges, resulting in poor outcomes. \cite{cakic2023developing}
Using COLMAP as an example, point cloud reconstruction relies on extracting and matching local features. However, since plant roots are underground and lack sufficient texture and color information, their complex structure and significant occlusion lead to incomplete reconstructions. Additionally, point cloud reconstructions often lack structured information and require further operations, such as skeleton extraction, to obtain useful phenotypic data. 
Deep learning-based bundle adjustment networks \cite{tang2018ba} struggle with the complex and intertwined structure of roots, making it difficult to find corresponding points across multiple views.
To address this, we propose a root skeleton extraction model that uses multi-view images to achieve accurate skeleton reconstruction. By employing a root detection network to locate lateral roots, triangulation and reprojection errors are used to determine their coordinates, connecting the starting points of lateral roots. Skeleton bundle adjustment is then applied to optimize the 3D skeleton and pose, ultimately reconstructing the main root skeleton. This approach overcomes the limitations of traditional methods, offering improved accuracy in root skeleton extraction.

\vspace{-1mm}
\section{Lateral Root Detection and Matching}
\vspace{-1mm}
\subsection{Lateral Root Detection}
\vspace{-1mm}
We have developed a method for precise extraction of the 3D skeleton of plant lateral roots. Due to the complexity and entanglement of root systems, directly matching the same lateral root in two images is challenging. To address this, we use an object detection network to identify detection boxes for lateral roots in each image. However, a single detection box may cover parts of multiple roots, meaning the same root could be partially covered by several boxes. Each detection box is ultimately interpreted as representing the most complete root it contains. By simultaneously outputting the starting and ending points of the roots, the network allows us to accurately identify the lateral roots and their directions, ensuring precise 3D skeleton extraction.


The upper part (Lateral root detection branch) of Fig.~\ref{arch}  illustrates the lateral root detection pipeline, employing a deep convolutional neural network $\mathcal{N}$ to process a root image $\mathbf{I}\in\mathbb{R}^{h\times w\times 3}$ into four output grids $\hat{\mathbf{G}} = \{\hat{\mathcal{G}}^s | s\in\{8, 16, 32, 64\}\}$, each representing object predictions $\hat{\mathbf{O}}$ with dimensions $\hat{\mathcal{G}}^s\in\mathbb{R}^{\frac{h}{s}\times \frac{w}{s}\times N_a \times N_o}$. 
 $ {N_a}$ includes $\text{anchor boxes } \mathbf{A}_s = (A_{w}, A_{h})$, and
$ {N_o}$ includes the objectness $\hat{p}_o$ (the probability that an object exists), $\text{anchor boxes } \mathbf{A}_s = (A_{w}, A_{h})$, the intermediate bounding box $\hat{t'} = (\hat{t'}_x, \hat{t'}_y, \hat{t'}_w, \hat{t'}_h)$, the intermediate keypoints $\hat{v}' = \{(\hat{v}'_{xk}, \hat{v}'_{yk})\}_{k=1}^{K}$ for the root. $\mathcal{N}$ is a YOLOv8\cite{Jocher_Ultralytics_YOLO_2023} feature extractor. The intermediate bounding box $\hat{t} = (\hat{t}_x, \hat{t}_y, \hat{t}_w, \hat{t}_h)$ of an object is predicted using grid coordinates, relative to the origin of the grid cell $(i, j)$:
\vspace{-1mm}
\begin{equation}
\begin{array}{l}
\widehat{t}_x = 2\sigma \left( \widehat{t}_x' \right) - 0.5, \quad \widehat{t}_y = 2\sigma \left( \widehat{t}_y' \right) - 0.5 \\
\widehat{t}_w = \frac{A_w}{s} \left( 2\sigma \left( \widehat{t}_w' \right) \right)^2, \quad \widehat{t}_h = \frac{A_h}{s} \left( 2\sigma \left( \widehat{t}_h' \right) \right)^2
\end{array}
\end{equation}

Keypoints $\hat{v}$ are predicted in the grid coordinates and relative to the grid cell origin $(i, j)$ using:
\vspace{-2mm}
\begin{small}
\begin{equation}
\widehat{v}_{xk} = \frac{A_w}{s} \left( 4\sigma \left( \widehat{v}_{xk}' \right) - 2 \right), \quad \widehat{v}_{yk} = \frac{A_w}{s} \left( 4\sigma \left( \widehat{v}_{yk}' \right) - 2 \right)
\end{equation}
\end{small}
The sigmoid function $\sigma$ facilitates learning by constraining the ranges of the object properties ( the x and y coordinates of the keypoints \(\hat{v}\) are \(\widehat{v}_{xk}\) and \(\widehat{v}_{yk}\). $\hat{v}_x^k$ and $\hat{v}_y^k$ are constrained to \(\pm \frac{2A_w}{s}\) and \(\pm \frac{2A_h}{s}\), respectively).
A target grid set $\mathbf{G}$ is established, utilizing a multi-task loss $\mathcal{L}(\hat{\mathbf{G}}, \mathbf{G})$ to train on object presence ($\mathcal{L}_{obj}$), intermediate bounding boxes ($\mathcal{L}_{box}$), and keypoints ($\mathcal{L}_{kps}$). The calculation of these loss components is as follows:
\vspace{-2.5mm}
\begin{equation}
    \mathcal{L}_{obj} = \sum_s \frac{\omega_s}{n(G^s)}\sum_{G^s}\mathrm{BCE}(\hat{p}_o, p_o \cdot \mathrm{CIoU}(\hat{\mathbf{t}}, \mathbf{t}))
\end{equation}
\vspace{-2mm}
\begin{equation}
    \mathcal{L}_{box} = \sum_s \frac{1}{n(\mathcal{O} \in G^s)}\sum_{\mathcal{O} \in G^s}1 - \mathrm{CIoU}(\hat{\mathbf{t}}, \mathbf{t})
\end{equation}
\vspace{-2mm}
\begin{small}
\begin{equation}
    \mathcal{L}_{kps} = \sum_s \frac{1}{n(\mathcal{O}^p \in G^s)}\sum_{\mathcal{O}^p \in G^s} \sum_{k=1}^K \delta(\nu_k > 0)\|\hat{\mathbf{v}}_k - \mathbf{v}_k\|_2
    \vspace{-1mm}
\end{equation}
\end{small}
Here, $\omega_s$ represents the weighting factor for the grid, $\mathrm{BCE}$ denotes the binary cross-entropy loss, CIoU is the
 complete intersection over union (CIoU) \cite{zheng2020distance}, and $\nu_k$ are indicators for the visibility of keypoints. During implementation, these losses are computed across a batch of images, employing grids that are organized in batches. The overall loss, denoted by $\mathcal{L}$, is calculated as the sum of the individual loss components, each multiplied by their respective weight, and then this sum is normalized by the batch size $N_b$:
 \vspace{-2mm}
\begin{equation}
\mathcal{L} = N_b(\lambda_{obj}\mathcal{L}_{obj} + \lambda_{box}\mathcal{L}_{box}  + \lambda_{kps}\mathcal{L}_{kps})
\vspace{-2mm}
\end{equation}

\subsection{Lateral Root Matching}
\label{boundingbox}
Once the lateral roots are detected, we will match the corresponding root branches in order to triagulate the 3D positions. Before triangulation, it is crucial to identify which bounding boxes in the pair of root images correspond to the same lateral root. We use a feature extraction module \cite{lindenberger2023lightglue} to extract keypoints from both root images, allowing us to associate bounding boxes through matched keypoints. Due to the intricate structure of roots and the lack of distinctive texture or color, keypoint matching outcomes can vary significantly. To address this, we implement a voting mechanism to facilitate reliable matching of lateral roots across the two images.

The system predicts the matching relationships between local features from root images $A$ and $B$ taken from different positions. It identifies detection boxes for the same lateral root by extracting and matching keypoints. Due to the complexity of root structures and the lack of texture and color information, keypoint matching is unstable, so a voting system is used to improve lateral root matching.

To find the corresponding bounding boxes in two images, we adopt an advanced strategy that involves constructing a \textit{Match Score Matrix} to systematically evaluate the correspondence between bounding boxes across pairs of images. The essence of this matrix is to assign quantitative scores to potential correspondences between bounding boxes in image 1 and image 2, leveraging keypoint matching. For each pair of matched keypoints $p1$ and $p2$ across image 1 and image 2, we update the score matrix $match[i,j]$, where $i$ and $j$ represent the bounding boxes in image 1 containing $p1$ and in image 2 containing $p2$, respectively. Each increment of $match[i, j]$ signifies a more credible matching relationship between these two bounding boxes. Upon completing the filling of the score matrix, we employ a greedy algorithm to select matching pairs, ensuring that each bounding box is paired with the best matching bounding box from the other image. During this process, we set a Matching Threshold that mandates a minimum number of matching keypoints between each pair of successfully matched bounding boxes to enhance the accuracy of the match.


\section{3D Skeleton Extraction}
With the corresponding root matching points, we are able to model the 3D roots, which we target at the 3D skeleton, which includes extensive root features, such as angles, distribution patterns, etc. 3D modeling consists of two parts: the 3D skeleton extraction of lateral roots and their connection forms in 3D space. Since we have chosen to model the lateral roots first, the modeling of the main root has adopted a method of connecting the starting points of the lateral roots to acquire the entire root system's topological structure (3D skeleton). 

\vspace{-1mm}
\subsection{Lateral Root 3D Skeleton Extraction}
In Section \ref{boundingbox}, we detected and matched lateral roots in two images, followed by triangulation to restore the 3D positions of the matched lateral root start and end points. Since the positions predicted by the network may have errors that affect triangulation accuracy, we performed pairwise triangulation on the start and end points across multiple images to obtain 3D coordinates and improve the reliability of skeleton extraction. For roots matched to the same bounding box in the second image, each point received two sets of reconstructed 3D coordinates $(x_1, y_1, z_1)$ and $(x_2, y_2, z_2)$, and the final coordinates were optimized via reprojection, giving more weight to points with smaller errors.
For the triangulation conducted between the first and second images, the resultant 3D coordinates are denoted as $\mathbf{X}_{i-1, i}$. Correspondingly, for the triangulation between the second and third images, the resultant coordinates are represented as $\mathbf{X}_{i, i+1}$. To enhance the precision of the ultimate 3D skeleton extraction, each set of coordinates is reprojected onto all three images (i-1, i, i+1), with the Euclidean distance errors from the reprojected points to the original 2D coordinates being calculated. $\text{error}_{i-1,i}$ represents the error of projecting $\mathbf{X}_{i-1,i}$ back onto all three images, while $\text{error}_{i, i+1}$ is the error associated with reprojecting $\mathbf{X}_{i,i+1}$ onto the third image.The triangulation results are weighted according to the error, with smaller error results given greater weight in determining the final 3D coordinates, optimizing the accuracy of skeleton extraction. By considering the cumulative projection errors of the two sets of triangulation results, a more precise 3D position estimation is achieved.
For lateral roots matched only in the second image, after 3D skeleton extraction, they are reprojected onto the third image. If they fall into the background or exceed a threshold distance from the root, the lateral root skeleton is removed. This completes the extraction of lateral root skeletons across different images.

\begin{figure}[t]
\begin{center}
\includegraphics[width=8cm, height=4cm]{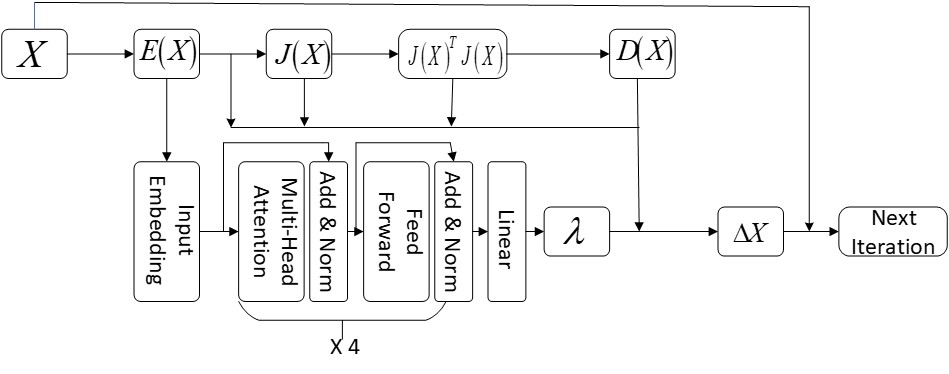}
\end{center}
\vspace{-8mm}
\caption{ The Structure of Skeleton Bundle Adjustment (SBA) Net }
\label{banet}
\vspace{-6mm}
\end{figure}
\begin{figure*}[t]
\begin{center}
\includegraphics[width=16cm, height=6cm]{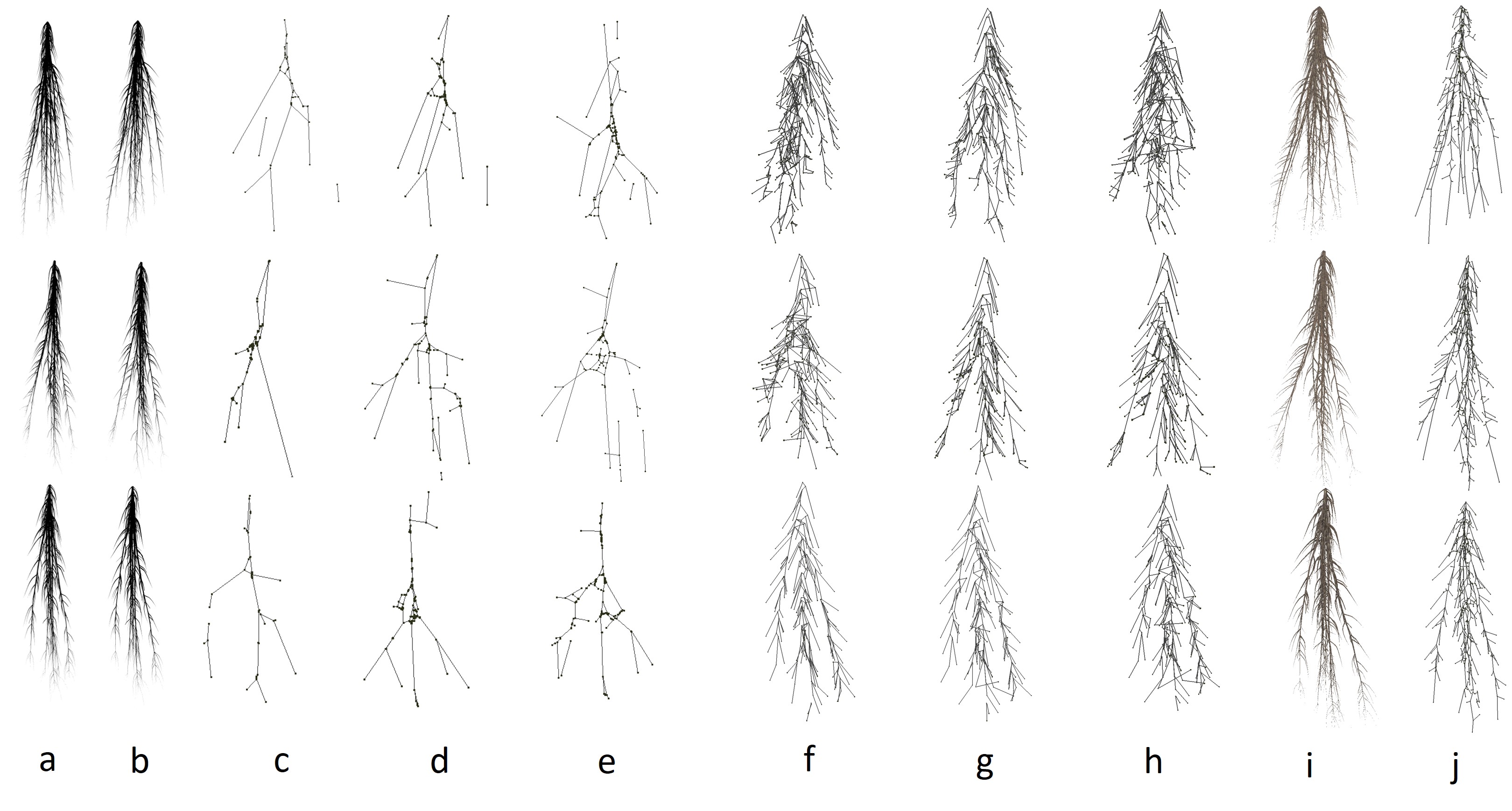}
\end{center}
\vspace{-8mm}
\caption{3D skeleton results: (a), (b) is different views, (c) is AdaBins \cite{bhat2021adabins}, (d) is MIM \cite{xie2023revealing}, (e) is Depth Anything \cite{yang2024depth}, (f), (g), (h) showcase the results of our 3D skeleton extraction from different perspectives.  (i), (j) are Ground Truth of root and skeleton.  }
\label{root_result}
\vspace{-7mm}
\end{figure*}

\vspace{-1mm}
\subsection{Skeleton Bundle Adjustment}
\vspace{-1mm}
Due to the inherent errors when calculating the 3D skeleton coordinates using small-angle multi-view images from different perspectives, we introduce skeleton bundle adjustment optimization into our network to mitigate these errors. Our target is to mitigate the reprojection error, simultaneously refining the camera poses $T = \left\{ {{T_i}|i = 1 \cdots {N_i}} \right\}$ and the coordinates of the 3D skeleton points $P = \left\{ {{p_j}|j = 1 \cdots {N_j}} \right\}$, thereby enhancing the accuracy of the 3D coordinates. The reprojection error is:
\vspace{-3mm}
\begin{equation}
\chi  = \arg \min \sum\limits_{i = 1}^{{N_i}} {\sum\limits_{j = 1}^{{N_j}} {\left\| {e_{i,j}^g\left( \chi  \right)} \right\|} } 
\end{equation}
\vspace{-3mm}
\begin{equation}
e_{i,j}^g\left( \chi  \right) = \pi \left( {{T_i},{p_i}} \right) - {q_{i,j}}
\vspace{-2mm}
\label{loss}
\end{equation}

$e_{i,j}^g\left( \chi  \right)$ measures the difference between a projected 3D coordinates and its corresponding feature point. The function $\pi $ projects 3D coordinates to image space, $q_{ij} = [x_{ij}, y_{ij}, 1]$ is the normalized homogeneous pixel coordinate, and $X = [T_1, T_2 \dots T_{N_i}, p_1, p_2 \dots p_{N_j}]$ contains all the points' and cameras' parameters.
We optimize the camera poses and the 3D skeletal structure by minimizing the reprojection error. Traditional Bundle Adjustment uses the Levenberg-Marquardt (LM) algorithm to optimize this error. However, the LM algorithm is non-differentiable due to its use of if-else decisions to update the damping factor and the number of iterations, making it impossible to perform feature learning through back-propagation. To address this issue, we fix the number of iterations and use a neural network to predict the damping factor. Our Skeleton Bundle Adjustment net (SBANet) is shown in Fig. \ref{banet}. We use the root feature points, root skeletal structure, and camera poses as inputs to the SBA layer. We use a 4-layer multi-head attention mechanism and a final 128-unit fully connected layer to fit the parameters.
During the forward pass, we compute the solution update \(X\) from the current solution \(X\) as follows: Using Eq. \ref{loss}, we calculate the error $E(X) = [e^f_{11}(X), e^f_{12}(X), \ldots, e^f_{N_i N_j}(X)]$ on all \(N_i\) images and \(N_j\) feature points, where \(X\) is the solution from the previous iteration. Then, we compute the Jacobian matrix \(J(X)\), the Hessian matrix \(J(X)^T J(X)\), and its diagonal matrix \(D(X)\). To predict the damping factor \(\lambda\), we use Multi-head attention layer to extract the value \(E(X)\) over all key points, obtaining a 128-dimensional feature vector, which is then fed into a fully connected layer to predict \(\lambda\). The update $\Delta X$ to the current solution according to Equation \ref{lm}.
\vspace{-3mm}
\begin{equation}
    \Delta X = {\left( {J{{\left( X \right)}^T}J\left( X \right) + \lambda D\left( X \right)} \right)^{ - 1}}J{\left( X \right)^T}E\left( X \right)
    \vspace{-2mm}
\label{lm}
\end{equation}

We use skeletal angles as the loss functions to train SBAnet. Skeleton angle loss: Since the angles between roots are consistent in real 3D space, we can use these angles as constraints during the reprojection process. For intersecting roots, we use the difference between the original angle and the new angle after projection as the loss function.

\vspace{-1mm}
\subsection{Main Root 3D Skeleton Extraction}
\vspace{-1mm}
For the connection of starting points of extracted lateral root skeletons, we simulate the growth process of plant roots to precisely extract the junction points of lateral roots.

\textbf{Reprojection:} The algorithm reprojects the extracted lateral root skeleton back to the original image to mimic the natural characteristic of plant root systems growing continuously downward from the top. This ensures the connection order of lateral roots aligns with the natural growth process.
    
\textbf{Matrix Initialization:} A matrix $M$ of the same size as the original image is defined to record the connection information of lateral root starting points. The points in the matrix representing the foreground (i.e., where the lateral roots are located) are initialized to $-1$, and the background points are initialized to $\infty$. The value of the pixel $p$ at the lateral root starting point is initialized to the index $i$ of that lateral root in the input list.
    
\textbf{Row-wise Updating:} Starting from the top of the image (i.e., the smallest $y$ value), the matrix $M$ is updated row by row for the values of foreground points. For each row, a continuous sequence of foreground points $[x_1, x_2]$ is considered, and the value update is based on the corresponding area $[x_1, x_2]$ of the previous row $y-1$.
    
\textbf{Value Update Rule:}
1. Count the values of all foreground points (non-$\infty$ values) in the area $[x_1, x_2]$ of the previous row and find the most frequent value $m$, indicating that this area belongs to the root with index $m$.
2. If all values in the sequence $[x_1, x_2]$ of the current row $y$ are $-1$, then update them all to $m$.
3. If there are non-$-1$ numbers $k$ in the sequence $[x_1, x_2]$ (indicating a new lateral root starting point), then update all points in this area to $k$.

 \textbf{Connection Recording:} Simultaneously record the current $k$ value and the $m$ value propagated from the previous row, representing a connection with the form $(m, k)$. After the root has fully grown, this represents the connection method of lateral roots on the image.
    
\textbf{Connection Retention:} In three-dimensional space, after connecting roots from the same list, only the connections that appear at least twice are retained $(m, n)$, indicating that $root[m]$ and $root[n]$ are adjacent lateral roots on the same main root. By connecting these lateral root starting points sequentially, the skeleton extraction of the main root skeleton is completed. Let $M_{ij}$ represent the points in the initialization matrix, where $i$ and $j$ represent the row and column coordinates of pixel. The update rule expressed as:
\vspace{-3mm}
\begin{equation}
M_{ij} = 
\begin{cases} 
m & \text{if } M_{i-1,j} = m \text{ for any } j \in [x_1, x_2] \\
& \quad \text{and } M_{ij} = -1\\
k & \text{if } M_{ij} \neq -1 \text{ and } j \in [x_1, x_2]\\
M_{ij} & \text{otherwise}
\vspace{-2mm}
\end{cases}
\end{equation}

where $m$ is the mode (the most frequently occurring value) of all foreground points (non-$\infty$ values) in the interval $[x_1, x_2]$ of the previous row $y-1$.
\vspace{-1mm}
\begin{small}
\begin{equation}
m = \underset{v}{\mathrm{argmax}}\, \left|\{M_{(y-1)j} = v \mid j \in [x_1, x_2] \text{ and } v \neq \infty\}\right|
\vspace{-2mm}
\end{equation}
\end{small}
$k$ is the index value of a new lateral root starting point, defined when there is at least one $M_{ij} \neq -1$ in the sequence $[x_1, x_2]$ of the current row $y$. $k$ is any value from the set of non-$-1$ values in that row (since $k$ represents a new starting point, it should be unique; if there are multiple different values of $k$, they should correspond to the same lateral root starting point). The calculation of $m$ reflects how we decide which root the current point belongs to based on data from the previous row, while the application of $k$ identifies a new starting point, indicating the direct assignment of lateral root starting points. Through these steps, the algorithm accurately simulates the natural growth process of plant root systems, achieving effective connection of lateral roots and extraction of the main root skeleton in digital images.

\vspace{-1mm}
\section{EXPERIMENTS}
\vspace{-2mm}
\subsection{Datasets}
\vspace{-1mm}
Due to the lack of publicly available datasets with 3D point clouds of plant roots and phenotyping-related annotations, we created a custom dataset for this study. Given the difficulty of directly comparing real root structures with algorithm-generated results, converting them into clean 3D mesh models enables more rigorous and quantitative evaluation. We captured 3D root point clouds using a high‑resolution scanner and performed manual denoising before annotation. The dataset includes 400 sweet potato root models, each containing approximately 200–400 fine root branches.  For each root system, we generated detailed mesh models along with 3 to 10 projection images from various angles by adjusting camera positions. Forty root systems were randomly selected as the test set.
\vspace{-1mm}
\subsection{Implementation Detail}
\vspace{-1mm}
Our network is composed of three modules: a target detection module, a feature point extraction module, and a Bundle Adjustment (BA) module. We train these modules separately, followed by overall fine-tuning. However, since we use a voting mechanism in the feature point extraction module to match lateral roots across views, this step is non-differentiable. Therefore, during fine-tuning, we freeze the feature point extraction module and only adjust the target detection and BA modules.

\begin{figure}[t]
\begin{center}
\includegraphics[width=8cm, height=7cm]{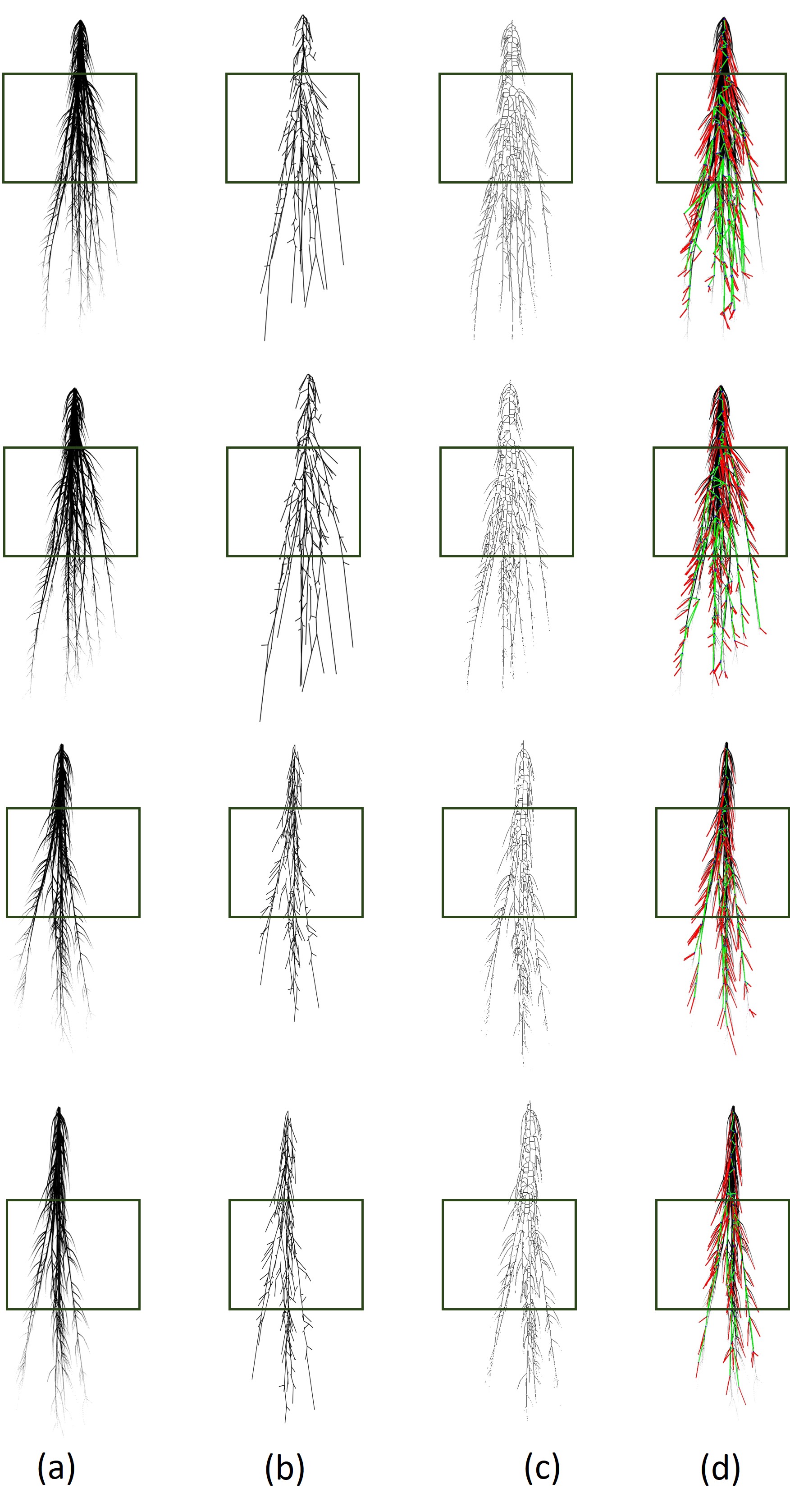}
\end{center}
\vspace{-6mm}
\caption{3D skeleton results: (a) is input. (b) is Ground Truth. (c) is Plant 3D \cite{niknejad2023phenotyping}. (d) is our result. }
\label{toolbox_result}
\vspace{-4mm}
\end{figure}

\begin{table}[]
\caption{Comparison of Precision and Recall for Lateral Root Matching vs. Direct keypoints detection by LightGlue\cite{lindenberger2023lightglue} and heatmap by SuperPoint\cite{detone2018superpoint}.}
\vspace{-3mm}
\resizebox{0.49\textwidth}{!}{
\vspace{-5mm}
\begin{tabular}{l|lll}
\hline
& \begin{tabular}[c]{@{}l@{}}\makecell{Keypoint  \\Detection}\end{tabular} & Heatmap & \begin{tabular}[c]{@{}l@{}}\makecell{Lateral Root \\Detection}\end{tabular} \\ \hline
\begin{tabular}[c]{@{}l@{}}Average Precision Of Lateral Root (2D)\end{tabular} & 0.72                                                          & 0.59    & \textbf{0.81}                                                              \\
\begin{tabular}[c]{@{}l@{}}Average Recall Of Lateral Root (2D)\end{tabular}    & 0.53                                                          & 0.49    & \textbf{0.69}                                                              \\ \hline

\end{tabular}}
\centering
\label{table2D}
\vspace{-6mm}
\end{table}

\vspace{-1mm}
\subsection{Lateral Root Matching and 3D Skeleton Extraction}
\vspace{-1mm}
Our algorithm differs from traditional direct point detection and point-to-point matching methods by prioritizing the detection, matching, and extraction of lateral root skeletons. Table \ref{table2D} provides a quantitative evaluation of our method, comparing the precision and recall of lateral root detection with those obtained using object detection or heatmap techniques to identify the starting and ending points of lateral roots. Table \ref{table3D} further compares the performance of using LightGlue versus direct feature point matching for point cloud reconstruction and skeleton extraction, highlighting the advantages of our algorithm in identifying and extracting complex root systems in 3D space.
Focusing on lateral roots reflects a deeper understanding of root system architecture, making detection and reconstruction more aligned with the intricate morphology of root systems. The results demonstrate that our method overcomes the limitations of traditional full-image analysis, enhancing the model’s accuracy and completeness through LightGlue point matching. Our approach excels in generating accurate and comprehensive root models, significantly improving precision and recall in both 2D and 3D contexts.

\begin{table}[t]
\centering
\caption{Comparison of Precision and Recall Metrics for Root Detection Using Lateral Root Matching Versus direct point matching by LightGlue\cite{lindenberger2023lightglue}.}
\vspace{-3mm}
\resizebox{0.49\textwidth}{!}{
\begin{tabular}{l|ll}
\hline
                                                                                 & Point Matching & Lateral Root Matching \\ \hline
\begin{tabular}[c]{@{}l@{}}Average Precision Of Lateral Root (3D)\end{tabular} & 0.57           & \textbf{0.77}                 \\
\begin{tabular}[c]{@{}l@{}}Average Recall  Of Lateral Root (3D)\end{tabular}    & 0.24           & \textbf{0.58}                  \\ \hline
\end{tabular}}
\label{table3D}
\end{table}

\vspace{-1mm}
\subsection{Main Root Connection}
\vspace{-1mm}
Figure \ref{root_result} and Table \ref{table_compare} present the results and performance of 3D skeleton extraction using different 3D reconstruction algorithms. The far left shows the input images for each experiment as a baseline for comparison. We compared the performance of supervised depth estimation algorithms (such as AdaBins) in predicting multi-view depth values and generating complete point cloud models, using Pc-Skeletor \cite{meyer2023cherrypicker} to extract 3D skeletons. However, traditional algorithms produced skeletons missing many root structures due to the complexity of root morphology, narrow depth range, and limited texture and color information, causing depth predictions to default to a flat plane. In contrast, our algorithm's 3D skeleton preserves more root structures and exhibits reasonable, consistent depth perception from multiple angles, indicating a 3D skeleton model with plausible depth. The far right shows ground truth models for comparison, further highlighting the superiority of our algorithm.

\begin{table}[]
\caption{Comparison of Precision and Recall Metrics for Root Detection Using Our Method Versus Other State-of-the-art Methods.}
\vspace{-3mm}
\resizebox{0.49\textwidth}{!}{
\vspace{-5mm}
\begin{tabular}{l|llll}
\hline
                      &\begin{tabular}[c]{@{}l@{}}AdaBins  \cite{bhat2021adabins} \end{tabular} &  
                           \begin{tabular}[c]{@{}l@{}}MIM  \cite{xie2023revealing} \end{tabular}
                       & \begin{tabular}[c]{@{}l@{}}Depth Anything\cite{yang2024depth}\end{tabular} & Ours \\ \hline
\begin{tabular}[c]{@{}l@{}}\makecell{Average Precision of \\Lateral Root (3D)}\end{tabular} & 0.31 & 0.33 & 0.37    &\textbf{0.77}                  \\
\begin{tabular}[c]{@{}l@{}}\makecell{Average Recall of \\Lateral Root (3D)}\end{tabular}    & 0.02 & 0.05 & 0.10           & \textbf{0.58 }                 \\ \hline
\end{tabular}}
\centering

\label{table_compare}
\vspace{-7mm}
\end{table}

\vspace{-1mm}
\subsection{Comparison with Other Skeleton Extraction Methods}
\vspace{-1mm}
Figure \ref{toolbox_result} compares the root skeletons reprojected to the original 2D image by our algorithm, the input images, the reprojected 3D skeletons extracted from these input images, and the 2D skeletons directly extracted using the Plant 3D Toolkit. Our method excels in maintaining the integrity of fine and short roots and accurately capturing the root system's topology, especially in areas with significant occlusion. This is crucial in root analysis, as occlusion can greatly affect the accurate representation of root connectivity and overall structure. Additionally, our method distinguishes between primary roots (green) and lateral roots (red), which is essential for understanding root structure and function, as these roots play different roles in plant growth and nutrient uptake. This distinction further highlights the advantages of our method over the Plant 3D Toolkit.
\vspace{-2mm}
\section{CONCLUSION}
\vspace{-1mm}
This paper presents a novel method for accurately extracting the 3D skeleton of plant root systems from limited viewpoints. The method successfully distinguishes and extracts both primary and lateral root skeletons, providing comprehensive, hierarchical structural information. It effectively handles complex multi-view inputs and 3D structure extraction, with potential applications in plant studies, root system model reconstruction using rendering algorithms, and automated breeding robots. By enabling precise root structure analysis, this method can be integrated into breeding robots to automatically select plants with optimal root systems, enhancing breeding efficiency and reducing manual labor in agricultural practices.

\textbf{Acknowledgment}: This work is supported by USDA NIFA grant No. 2021-67021-34199 and NSF Awards NO. 2340882, 2334624, 2334246, and 2334690.


\vspace{-2mm}
{
\bibliographystyle{ieee_fullname}
\bibliography{egbib}
}

\end{document}